\documentclass[runningheads]{llncs}
\usepackage[T1]{fontenc}
\usepackage{graphicx}
\usepackage{comment}
\usepackage{tikz}
\usepackage[colorlinks=true, linkcolor=blue, citecolor=blue, urlcolor=blue]{hyperref}
\usepackage{color}

\usepackage{cite}
\usepackage{amsmath,amssymb,amsfonts}
\usepackage{algorithmic}
\usepackage{graphicx}
\usepackage{textcomp}
\usepackage{tabularx}
\usepackage{xcolor}
\usepackage{soul}
\usepackage{enumitem}
\usepackage{placeins}  

\def\BibTeX{{\rm B\kern-.05em{\sc i\kern-.025em b}\kern-.08em
    T\kern-.1667em\lower.7ex\hbox{E}\kern-.125emX}}
\makeatletter
\newcommand{\printfnsymbol}[1]{%
  \textsuperscript{\@fnsymbol{#1}}%
}
\makeatother
\begin{document}
\title{CATVis: Context-Aware Thought Visualization}

%
%

\author{
Tariq Mehmood\inst{1}\thanks{These authors contributed equally to this work.}\and
Hamza Ahmad\inst{2}\printfnsymbol{1}\and
Muhammad Haroon Shakeel\inst{3}\and
Murtaza Taj\inst{1}
}

\authorrunning{T. Mehmood et al.}

\institute{
Lahore University of Management Sciences, Lahore, Pakistan\\
\email{\{23030009, murtaza.taj\}@lums.edu.pk}
\and
Forman Christian College, University, Lahore, Pakistan\\
\email{241548068@formanite.fccollege.edu.pk}
\and
Arbisoft, Pakistan\\
\email{mhfateen@gmail.com}
}

    
%
\maketitle              
\begin{abstract}
EEG-based brain-computer interfaces (BCIs) have shown promise in various applications, such as motor imagery and cognitive state monitoring. However, decoding visual representations from EEG signals remains a significant challenge due to their complex and noisy nature. We thus propose a novel 5-stage framework for decoding visual representations from EEG signals: (1) an EEG encoder for concept classification, (2) cross-modal alignment of EEG and text embeddings in CLIP feature space, (3) caption refinement via re-ranking, (4) weighted interpolation of concept and caption embeddings for richer semantics, and (5) image generation using a pre-trained Stable Diffusion model. We enable context-aware EEG-to-image generation through cross-modal alignment and re-ranking. Experimental results demonstrate that our method generates high-quality images aligned with visual stimuli, outperforming SOTA approaches by $13.43\%$ in Classification Accuracy, $15.21\%$ in Generation Accuracy and reducing Fréchet Inception Distance by 
$36.61\%$, indicating superior semantic alignment and image quality.

\end{abstract}
\section{Introduction}

Brain-computer interfaces (BCIs) decode neural signals to enable communication, particularly for individuals with severe neuromotor impairments. Electroencephalography (EEG), a non-invasive and affordable technique, is widely used in BCIs due to its high temporal resolution, despite challenges such as low spatial resolution \cite{varbu2022past}, low signal-to-noise ratio \cite{chaddad2023electroencephalography}, and artifact contamination \cite{8369420}. Decoding visual representations from EEG remains difficult, as traditional feature-based approaches struggled with limited datasets, while recent deep learning advances have improved reconstruction fidelity \cite{9515997, 9430619}.

Existing approaches like ThoughtViz \cite{tirupattur2018thoughtviz} and Brain2Image \cite{kavasidis2017brain2image} primarily focus on visualizing the conceptual information captured by neural signals, often overlooking the contextual elements, such as color, setting, or interaction. By neglecting these finer details, current methods provide only a coarse approximation of the visual stimulus. To address this gap, we propose a framework that not only improves the concept recognition but also captures the finer details by incorporating textual descriptions as an intermediary representation.


Our framework employs the EEG–image pairs dataset from \cite{palazzo2020decoding, spampinato2017deep}, in which each EEG sample is paired with a corresponding visual stimulus. We begin by generating a caption for each stimulus to obtain a ground-truth description that serves as the basis for alignment in the CLIP \cite{radford2021learning} feature space. Next, our encoder classifies the EEG signals into conceptual categories, and we interpolate the concept and caption embeddings to form a semantically rich conditioning vector for Stable Diffusion (SD), enabling the generation of high-fidelity images.
%
%
Our end-to-end approach leverages state-of-the-art techniques in natural language processing and computer vision to create a complete pipeline from neural activity to visual output. Our contributions can be summarized as follows:

\begin{itemize}[label=\textbullet]
    \item Unlike prior work \cite{tirupattur2018thoughtviz, kavasidis2017brain2image}, which focused on conceptual decoding (class labels), we enable context-aware (descriptive information) EEG-to-image generation through cross-modal alignment and re-ranking, boosting generation accuracy by 7.51\% over baselines.
    \item Unlike resource-intensive self-supervised methods \cite{bai2023dreamdiffusion, fu2023brainvis}, we achieve a 13.43\% higher classification accuracy using a supervised EEG Conformer \cite{9991178eeg-conformer}, demonstrating that robust EEG representations can be learned without extensive pretraining.
    \item We modulate the emphasis on concept vs. context. This increases the semantic fidelity of reconstructed images, aligning them more closely with the true visual content.
\end{itemize}

\section{Related Work}

Decoding visual information from brain activity and reconstructing images has gained significant attention in recent research. Initial work in this domain leveraged the discriminative power of GANs to directly generate realistic images from brain activity \cite{kavasidis2017brain2image, tirupattur2018thoughtviz, singh2023eeg2image, mishra2023neurogan}. Brain2Image \cite{kavasidis2017brain2image} used an LSTM to classify EEG into 40 ImageNet classes, conditioning VAE and GAN generation with discriminative features. ThoughtViz \cite{tirupattur2018thoughtviz} built on this with 1D/2D-CNNs to classify EEG into object classes, driving GAN synthesis. Yet, both \cite{kavasidis2017brain2image, tirupattur2018thoughtviz} rely on class-discriminative features, limiting their capacity to capture contextual details beyond basic conceptual decoding, thus reducing their utility for detailed visual reconstruction.

Diffusion models, inspired by non-equilibrium thermodynamics, have emerged as an effective generative modeling approach \cite{rombach2022high}. 
DreamDiffusion \cite{bai2023dreamdiffusion} fine-tunes Stable Diffusion \cite{rombach2022high} by conditioning on robust EEG features obtained through a Masked Autoencoder approach. These features are further aligned with CLIP image embeddings to improve cross-modal consistency, enabling high-quality image generation. However, DreamDiffusion relies on computationally intensive self-supervised pretraining, which limits its scalability and accessibility, particularly for resource-constrained settings. 


In contrast, BrainVis \cite{fu2023brainvis} removes the need for external data by training a robust EEG encoder with Latent Masked Modeling, leveraging both time-domain and frequency-domain features while incorporating reconstruction and classification objectives. These features are aligned with interpolated semantic embeddings (coarse-grained labels and fine-grained captions). The aligned embeddings are used as conditioning for cascaded diffusion models, resulting in better semantic accuracy of generated images.  However, BrainVis’ complex cascaded diffusion architecture increases computational overhead.

\section{Methodology}

Our current methodology comprises five main components: (1) Conformer-Based EEG Encoder for concept classification; (2) Cross-modal alignment in CLIP feature space between EEG and caption embeddings; (3) Caption refinement through re-ranking of retrieved embeddings from CLIP feature space; (4) Weighted interpolation of concept and caption for semantically rich embedding; (5) Image generation using pretrained Stable Diffusion.

\begin{figure}[h]
    \centering
    \includegraphics[width=0.8\textwidth]{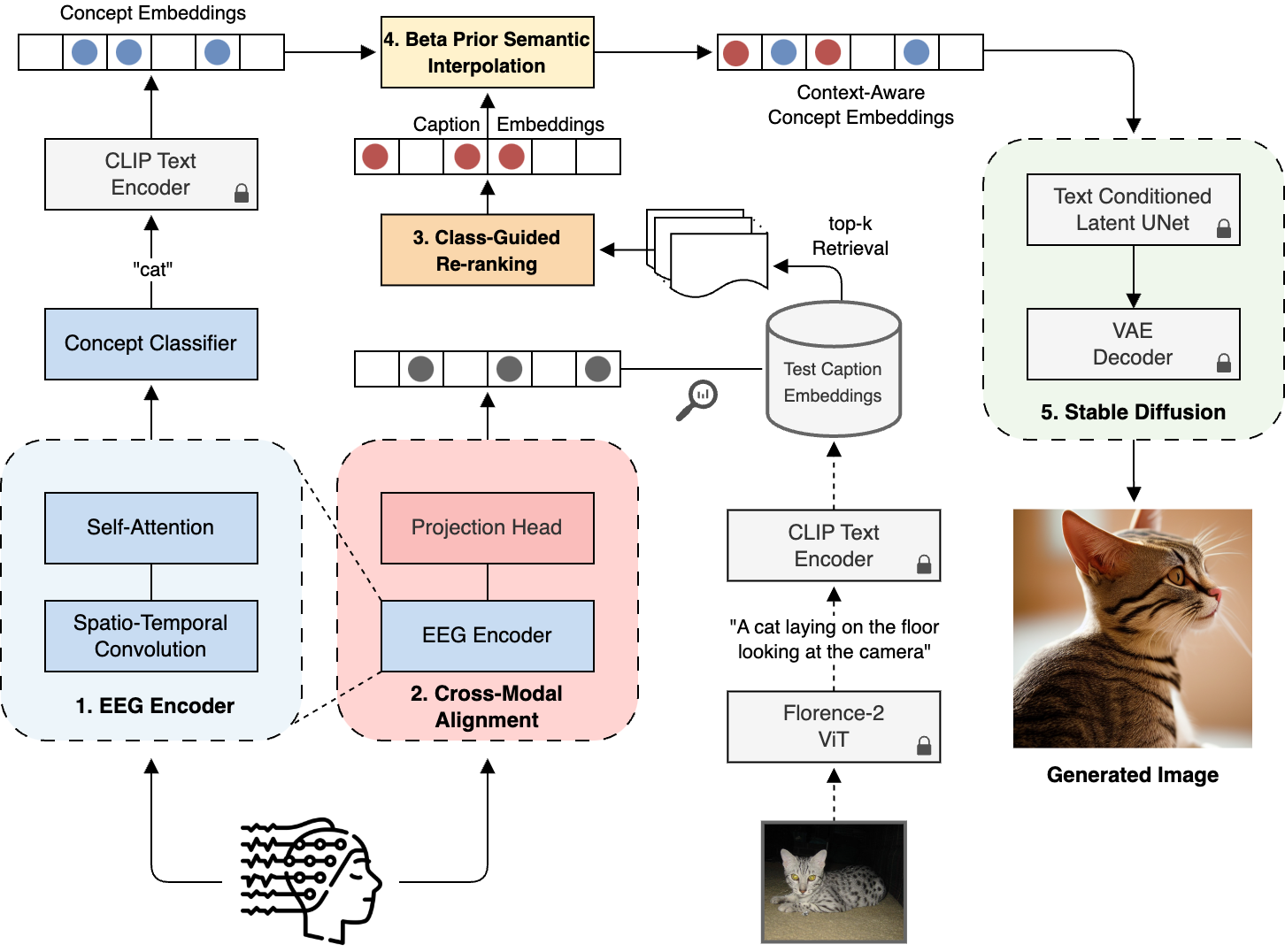} 
    \caption{Overview of our proposed five-stage framework for reconstructing images from EEG signals. The numbered modules correspond to the key steps in the methodology.}
    \label{fig:system_overview}
\end{figure}

\subsection{EEG Encoder for Concept Classification}


Unlike self-supervised approaches~\cite{bai2023dreamdiffusion, fu2023brainvis} that rely on self-supervised reconstruction objectives, we adopt a supervised EEG Conformer~\cite{9991178eeg-conformer} encoder, trained from scratch on our dataset. The architecture leverages spatio-temporal convolutions and a stacked Transformer encoder to capture EEG signals' local and long-term dependencies, with a fully connected classification head to predict class labels.



\subsection{Cross-modal Alignment}

\begin{figure}[h]
    \centering
    \includegraphics[width=0.85\textwidth]{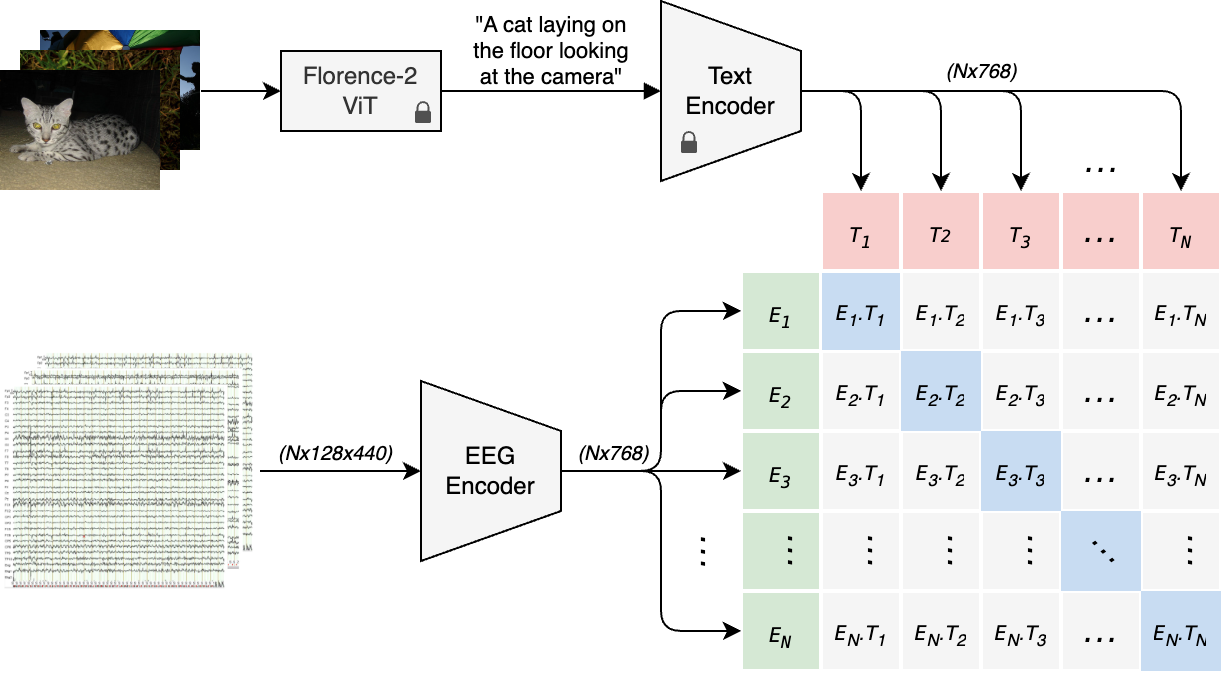} 
    \caption{Cross-modal alignment of EEG and text embeddings via contrastive loss. We project EEG signals and captions into a joint representation space, where positive pairs (blue cells) are brought closer, and negative pairs (gray cells) are pushed apart.}
    \label{fig:clip}
\end{figure}

Our cross-modal alignment framework learns to project EEG signals into CLIP's joint embedding space through contrastive learning. We adapt the EEG Conformer architecture by replacing its final classification layer with a linear projection to 768 dimensions followed by L2-normalization. The caption embeddings are obtained via CLIP's pretrained ViT-L/14 text encoder.
%
%
We optimize the following symmetric contrastive loss:
\begin{equation}
\label{eq:total-loss}
\mathcal{L} = \frac{1}{2}(\mathcal{L}{\text{EEG→Text}} + \mathcal{L}{\text{Text→EEG}})
\end{equation}
%
%
To be more explicit, we have two directional terms. For the EEG$\rightarrow$Text direction:
\begin{equation}
\label{eq:eeg2text}
\mathcal{L}_{\mathrm{EEG}\to\mathrm{Text}}
\;=\;
- \sum_{i=1}^{B} \log 
\Biggl(
  \frac{\exp\bigl(\mathbf{e}_i^{\mathrm{EEG}} \cdot \mathbf{e}_i^{\mathrm{text}} \,/\, \tau \bigr)}
       {\sum_{j=1}^{B}\exp\bigl(\mathbf{e}_i^{\mathrm{EEG}} \cdot \mathbf{e}_j^{\mathrm{text}} \,/\, \tau \bigr)}
\Biggr),
\end{equation}
where $\mathbf{e}_i^{\mathrm{EEG}}$ is the normalized embedding for the EEG sample $i$, $\mathbf{e}_j^{\mathrm{text}}$ is the embedding for the text sample $j$, $B$ is the mini-batch size, and $\tau$ is the temperature parameter.
Similarly, for the Text$\rightarrow$EEG direction:
\begin{equation}
\label{eq:text2eeg}
\mathcal{L}_{\mathrm{Text}\to\mathrm{EEG}}
\;=\;
- \sum_{i=1}^{B} \log 
\Biggl(
  \frac{\exp\bigl(\mathbf{e}_i^{\mathrm{text}} \cdot \mathbf{e}_i^{\mathrm{EEG}} \,/\, \tau \bigr)}
       {\sum_{j=1}^{B}\exp\bigl(\mathbf{e}_i^{\mathrm{text}} \cdot \mathbf{e}_j^{\mathrm{EEG}} \,/\, \tau \bigr)}
\Biggr).
\end{equation}
In this setup, each EEG embedding $\mathbf{e}_i^{\mathrm{EEG}}$ is ``pulled closer'' to its corresponding text embedding $\mathbf{e}_i^{\mathrm{text}}$ and pushed away from non-matching captions in the batch. Figure \ref{fig:clip} demonstrates this alignment process.

\subsection{Caption Refinement}

Through cross-modal alignment, we align EEG representations with CLIP embeddings, which contain fine-grained semantic information. However, because EEG signals reflect only the subject’s perception rather than every detail of the captioned image, they often fail to capture finer contextual cues. As a result, the neural response typically centers on the main concept and some contextual elements, leading to alignment inconsistencies when EEG embeddings are compared with CLIP text embeddings containing richer semantic information.

To address this limitation, we employ a two-stage retrieval process:

\begin{enumerate}
    \item \textbf{Initial Retrieval.} Given an EEG embedding 
    $\mathbf{e}^{\mathrm{EEG}} \in \mathcal{Q}$ (the \emph{query}) 
    and a collection of text embeddings 
    $\mathcal{X} = \{\mathbf{e}_{1}^{\mathrm{text}},\dots,\mathbf{e}_{n}^{\mathrm{text}}\}$ 
    of $n$ caption embeddings belonging to the entire corpus, 
    we retrieve the top-$k$ most similar captions in the CLIP feature space as:  
    \begin{equation}
    \label{eq:topk-argmax}
    \arg\max_{x \,\in\, \mathcal{X}}^{(k)} \cos\bigl(\mathbf{e}^{\mathrm{EEG}},\,\mathbf{e}^{\mathrm{text}}\bigr),
    \end{equation}

    This operation returns a subset of $\mathcal{X}$ that best match the neural response.

    \item \textbf{Class-Guided Re-ranking.} These top-\(k\) candidates are then re-ranked according to their similarity to the predicted CLIP-encoded class label obtained from our concept classifier, ensuring that the final caption aligns with the primary object most likely perceived by the subject. This re-ranking step yields 6.99\% higher Generation Accuracy (GA), 6.30\% higher Inception Score (IS), and 3.43\% lower Fréchet Inception Distance (FID) versus without re-ranking (see Table \ref{tab:ablation_reranking}).
\end{enumerate}

\subsection{Beta Prior Semantic Interpolation}





To balance object-focused conceptual and fine-grained descriptive information, we interpolate the predicted concept embedding from our EEG classifier with the re-ranked caption embedding. Let \(\mathbf{e}^{\text{class}} \in \mathbb{R}^{768}\) be the CLIP-encoded class label predicted by the EEG Conformer, and \(\mathbf{e}^{\text{text}} \in \mathbb{R}^{768}\) the top-ranked CLIP caption embedding. We sample \(\lambda \sim \mathrm{Beta}(\alpha, \beta)\) and compute:

\begin{equation}
\label{eq: interpolation}
  \mathbf{z} = \lambda \mathbf{e}^{\text{class}} + (1 - \lambda) \mathbf{e}^{\text{text}},
\end{equation}
where \(\mathbf{z}\) conditions the diffusion process.

\subsection{Stable Diffusion}

We use a pre-trained Stable Diffusion model \cite{rombach2022high} to transform the EEG-derived conditioning vector $\mathbf{z}$ into photorealistic images. Specifically, we condition the denoising process on $\mathbf{z}$ by integrating it into the cross-attention mechanism of Stable Diffusion’s UNet, guiding the reverse diffusion process to reconstruct the image latent. Finally, the VAE decoder converts the latent representation into the pixel space, yielding the generated image. By leveraging Stable Diffusion’s generative capabilities with EEG-driven embeddings, our approach aims to generate images that capture both the primary object class and finer contextual details inferred from neural activity.

\section{Experiments}

\subsection{Dataset}
We use the EEG-ImageNet dataset \cite{spampinato2017deep, palazzo2020decoding}, featuring EEG signals from 6 subjects exposed to $2000$ ImageNet images ($40$ concepts, $50$ images each) via Rapid Serial Visual Presentation, recorded with a 128-electrode headset. Each trial yields $12,000$ EEG-image pairs, with signals ($128$ channels, $440$ timepoints) preprocessed to retain the $20-460$ ms interval and band-pass filtered at $55-95$ Hz for cognitive relevance \cite{palazzo2020decoding}. We adopt the provided train/validation/test splits. Ground-truth captions are generated using Microsoft Florence-2 Large \cite{xiao2024florence} to pair EEG signals with image descriptions for cross-modal alignment.

\subsection{Implementation Details}
We implement our framework in PyTorch (Python 3.10), using the EEG Conformer \cite{9991178eeg-conformer} as the encoder backbone, modified to output 768-dimensional embeddings for CLIP alignment. The temporal convolution uses $k = 40$ kernels of size $(1, 25)$, while the spatial convolution applies the same number of kernels with a size of $(ch, 1)$, where $ch = 128$. The encoder is trained for 120 epochs with Adam ($\eta =  0.0001$) on cross-entropy loss. The cross-modal alignment module is trained separately for 100 epochs, minimizing InfoNCE loss (\(\tau = 0.07\)) with random sampling for mini-batches. Interpolation coefficients are sampled from a Beta distribution (\(\alpha = 10, \beta = 10\)). Image generation uses pre-trained Stable Diffusion v1-5 \cite{Rombach_2022_CVPR}, conditioned via CLIP ViT-L/14 embeddings.

\subsection{Evaluation Metrics}

\subsubsection{EEG Classification Accuracy (CA)}

CA measures the EEG encoder’s ability to extract discriminative latent features from EEG signals, with higher values indicating better class prediction accuracy.

\subsubsection{\textit{N}-way Top-\textit{K} Classification Accuracy of Generation (GA)}

Following \cite{chen2023seeing, bai2023dreamdiffusion, fu2023brainvis}, GA evaluates semantic correctness of generated images using an \textit{N}-way top-\textit{K} classification task. We use a pre-trained ImageNet1K classifier \cite{dosovitskiy2020vit} to compare ground-truth and generated images, selecting \textit{N}-1 random classes plus the ground-truth class, and checking if the top-\textit{K} prediction matches the ground truth over 50 trials. We set \textit{N} = 50 and \textit{K} = 1, consistent with prior evaluations.


\subsubsection{Inception Score (IS)}

IS \cite{salimans2016improved} assesses image quality and diversity using a pre-trained Inception-v3 classifier \cite{szegedy2015rethinking}. It measures classifier confidence and class distribution diversity, where higher scores reflect better quality and variety.

\subsubsection{Fréchet Inception Distance (FID)}

FID \cite{heusel2017gans} evaluates generation quality by computing the feature distribution distance between ground-truth and generated images using Inception-v3 \cite{szegedy2015rethinking}. Lower FID values indicate better alignment.



\section{Results \& Discussion}

\begin{figure}[h]
    \centering
    \includegraphics[width=0.85\textwidth]{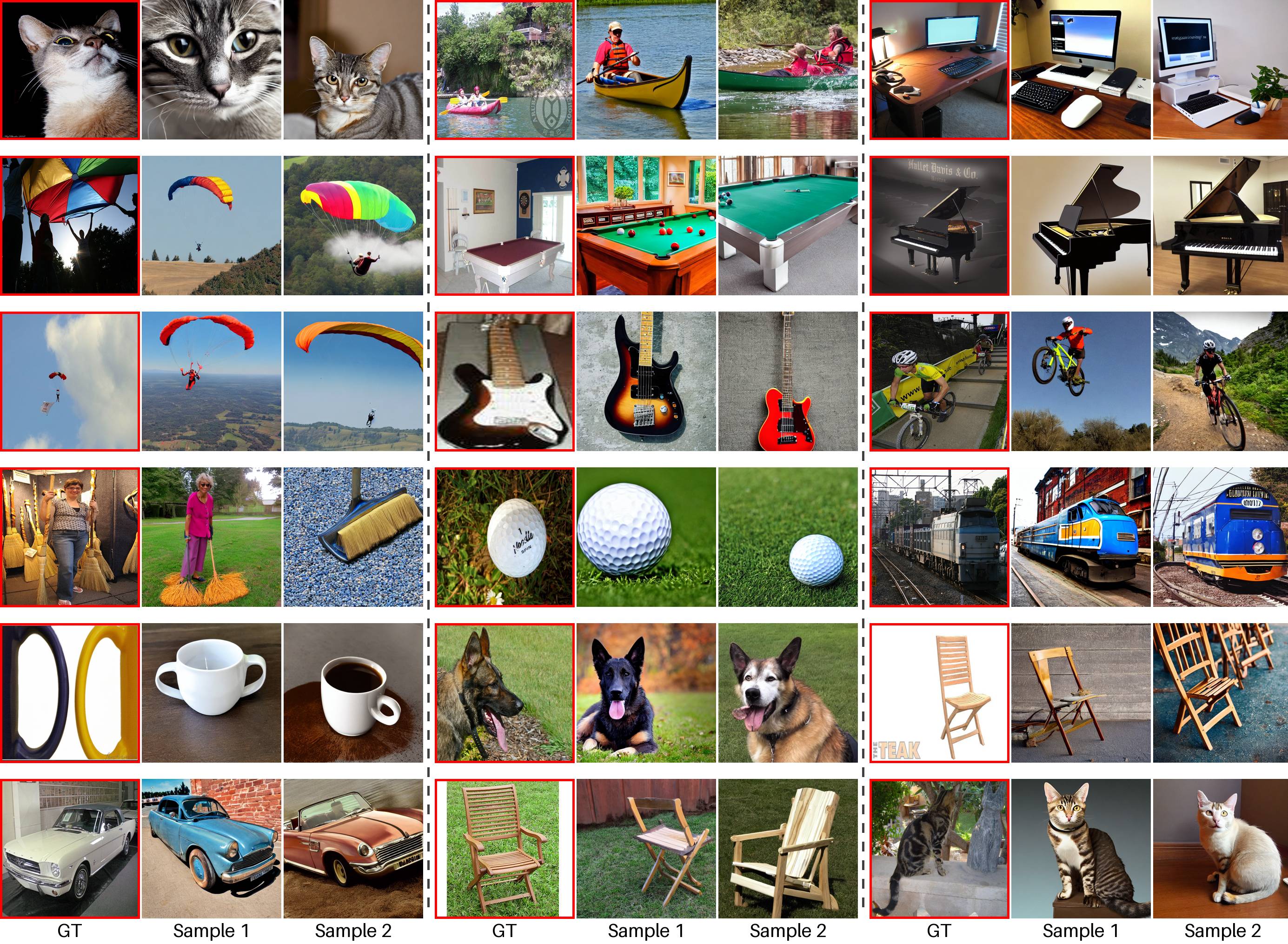} 
    \caption{Images generated from EEG signals using our framework. The images in red boxes are visual stimuli, followed by two corresponding generated samples.}
    \label{fig:gen_samples}
\end{figure}

Our framework demonstrates significant advancements in decoding visual representations from EEG signals, as evidenced by both quantitative metrics and qualitative outputs. Table \ref{tab:comp_classification} shows that our supervised EEG Conformer achieves a Top-1 CA of 61.09\%, surpassing BrainVis \cite{fu2023brainvis} by 13.43\%, highlighting the encoder’s ability to extract robust, discriminative features without resource-intensive pretraining. For image generation, our method achieves a GA of 0.5678 for Subject 4 (Table~\ref{tab:comp_generation_s4}) and 0.5265 on average (Table~\ref{tab:comp_generation}), outperforming DreamDiffusion by 23.91\% (Subject 4) and BrainVis by 15.21\% (average), alongside an IS of 37.4948 and a FID of 80.2949, reflecting a 36.61\% reduction in FID compared to BrainVis. These results underscore improved semantic alignment and image quality, as visually confirmed in Figure \ref{fig:gen_samples}, where generated images closely resemble their ground-truth stimuli in both conceptual and contextual details. Collectively, these findings validate the efficacy of our five-stage approach.

\subsection{Ablation Studies}

We ablated our EEG Conformer and interpolation strategy to assess their roles. Table~\ref{tab:ablation_encoder} shows removing the Transformer reduces Top-1 CA by 5.76\%, as it captures global temporal dependencies across EEG channels, vital for long-range neural patterns. Excluding the convolutional block drops CA by 45.77\%, since it extracts local spatio-temporal features essential for stimulus-specific responses. Table \ref{tab:ablation_interpolation} tests interpolation weights: a concept-heavy weight (0.75) yields the highest GA (61.48\%) by emphasizing class-specific accuracy from the EEG classifier, while a caption-heavy weight (0.75) lowers GA (53.93\%) due to potential noise in descriptive details. The Beta(10,10) scheme balances both, achieving the best IS (35.79) by enhancing diversity via captions. These results validate the synergy of encoder components and the interpolation’s role in balancing conceptual and contextual fidelity for Stable Diffusion conditioning.

\begin{table}[b]
    \centering
    \begin{minipage}[t]{.48\textwidth}
      \caption{Comparison of EEG classification accuracy with existing approaches.}
      \centering
      \footnotesize
      \setlength{\tabcolsep}{1pt}
      \begin{tabular}{l|cccc}
        \hline
        \textbf{Method} & \textbf{Top1}$\uparrow$ & \textbf{Top3} $\uparrow$ & \textbf{Top5}$ \uparrow$ & \textbf{F1} $\uparrow$ \\
        \hline
        Brain2Image & 0.16 & 0.38 & 0.55 & 0.16 \\
        KD-STFT \cite{ferrante2024decoding} & 0.41 & 0.75 & 0.88 & 0.40 \\
        BrainVis \cite{fu2023brainvis} & 0.48 & 0.79 & 0.91 & 0.43 \\
        CATVis (Ours) & \textbf{0.61} & \textbf{0.94} & \textbf{0.98} & \textbf{0.61} \\
        \hline
      \end{tabular}
      \label{tab:comp_classification}

        \caption{Effect of caption re-ranking on generation performance.}
        \centering
        \footnotesize
        \begin{tabular}{l|ccc}
        \hline
        \textbf{Method} & \textbf{GA} $\uparrow$ & \textbf{IS} $\uparrow$ & \textbf{FID} $\downarrow$ \\
        \hline
        w/o Re-ranking    &0.5307&32.4678&86.0763\\
        w/ Re-ranking     &\textbf{0.5678}&\textbf{34.5115}&\textbf{83.1264}\\
        \hline
        \end{tabular}
        \label{tab:ablation_reranking}
      
    \end{minipage}%
    \hfill
    \begin{minipage}[t]{.48\textwidth}
      \caption{Comparison with image reconstruction baselines on all Subjects.}
      \centering
      \footnotesize
      \setlength{\tabcolsep}{1pt}
        \begin{tabular}{l|ccc}
        \hline
        \textbf{Methods} & \textbf{GA} $\uparrow$ & \textbf{IS} $\uparrow$ & \textbf{FID} $\downarrow$ \\
        \hline
        Brain2Image \cite{kavasidis2017brain2image} & - & 5.07 & - \\
        ESG-ADA \cite{singh2024learning} & - & 10.82 & 174.13 \\
        BrainVis \cite{fu2023brainvis} & 0.46 & 31.00 & 126.66 \\
        CATVis (Ours) & \textbf{0.53} & \textbf{37.50} & \textbf{80.30} \\
        \hline
      \end{tabular}
      \label{tab:comp_generation}

        \caption{Comparison with image reconstruction baselines on Subject 4.}
        \centering
        \footnotesize
          \begin{tabular}{l|ccc}
        \hline
        \textbf{Methods} & \textbf{GA} $\uparrow$ & \textbf{IS} $\uparrow$ & \textbf{FID} $\downarrow$ \\
        \hline
        DreamDiffusion \cite{bai2023dreamdiffusion} & 0.46 & - & - \\
        BrainVis \cite{fu2023brainvis} & 0.49 & 31.53 & 121.02 \\
        CATVis (Ours) & \textbf{0.57} & \textbf{34.51} & \textbf{83.13} \\
        \hline
      \end{tabular}
      \label{tab:comp_generation_s4}
      
    \end{minipage}%
    
\end{table}

\FloatBarrier

\section{Conclusion}

Our five-stage framework advances EEG-based visual reconstruction, achieving a 13.43\% higher CA and 15.21\% improved GA over state-of-the-art methods, with a 36.61\% FID reduction. By integrating supervised EEG encoding, CLIP alignment, caption refinement, and interpolation, we capture both conceptual and contextual details, outperforming prior approaches without extensive pre-training. This work offers a scalable solution for BCI applications, with future potential in refining cross-modal alignment and exploring real-time visualization.


\begin{table}[ht]
    \centering
    \begin{minipage}[t]{.48\textwidth}
    \caption{Classification Accuracy (CA) of EEG Encoder w/wo CNN and w/wo Transformer (Trans.)}
    \centering
    \footnotesize
    \begin{tabular}{cc|cccc}
    \hline
    \textbf{CNN} & \textbf{Trans.} & \textbf{Top1} $\uparrow$ & \textbf{Top3} $\uparrow$ & \textbf{Top5} $\uparrow$ & \textbf{F1} $\uparrow$ \\
    \hline
    -  & \checkmark        & 15.32    & 34.50    & 48.27    & 0.16     \\
    \checkmark & -         & 55.33    & 90.84    & 97.70    & 0.55     \\
    \checkmark & \checkmark & \textbf{61.09} & \textbf{93.84} & \textbf{98.15} & \textbf{0.61} \\
    \hline
    \end{tabular}
    \label{tab:ablation_encoder}
    \end{minipage}
    \hfill
    \begin{minipage}[t]{.48\textwidth}
    \caption{Generation accuracy (GA) for concept–caption interpolation schemes, with weights $\omega_{\text{concept}}$ and $\omega_{\text{caption}}$.}
    \centering
    \footnotesize
    \begin{tabular}{cc|ccc}
    \hline
    $\omega_{\text{concept}}$ & $\omega_{\text{caption}}$ & \textbf{GA} $\uparrow$ & \textbf{IS} $\uparrow$ & \textbf{FID} $\downarrow$ \\
    \hline
    0.75 & 0.25  & \textbf{61.48} & 34.34 & \textbf{82.33} \\
    0.25 & 0.75  & 53.93 & 30.50 & 83.30 \\
    0.5 & 0.5 & 56.79 & 34.50 & 86.66 \\
    \multicolumn{2}{c|} {Beta(10,10)}       & 56.86 & \textbf{35.79} & 84.13 \\
    \hline
    \end{tabular}
    \label{tab:ablation_interpolation}
    \end{minipage}
\end{table}

\end{document}